\documentclass[10pt]{article} 
\usepackage[preprint]{tmlr}

\usepackage{amsmath,amsfonts,bm}

\def\eqref#1{equation~\ref{#1}}

\def\1{\bm{1}}

\DeclareMathAlphabet{\mathsfit}{\encodingdefault}{\sfdefault}{m}{sl}
\SetMathAlphabet{\mathsfit}{bold}{\encodingdefault}{\sfdefault}{bx}{n}

\usepackage{hyperref}
\usepackage{url}

\usepackage{algorithm}
\usepackage{algpseudocode}
\usepackage{graphicx}

\title{Solving Sokoban using Hierarchical Reinforcement Learning with Landmarks}

\author{Sergey Pastukhov    \\
  Amsterdam \\
  \texttt{omikad@gmail.com}                
}

\def\month{MM}  
\def\year{YYYY} 
\def\openreview{\url{https://openreview.net/forum?id=XXXX}} 

\begin{document}

\maketitle

\begin{abstract}
We introduce a novel hierarchical reinforcement learning (HRL) framework that performs top-down recursive planning via learned subgoals, successfully applied to the complex combinatorial puzzle game Sokoban. Our approach constructs a six-level policy hierarchy, where each higher-level policy generates subgoals for the level below. All subgoals and policies are learned end-to-end from scratch, without any domain knowledge, demonstrations, or manually defined abstractions. Our results show that the agent can generate long action sequences from a single high-level call. While prior work has explored 2–3 level hierarchies and subgoal-based planning heuristics, we demonstrate that deep recursive goal decomposition can emerge purely from learning, and that such hierarchies can scale effectively to hard puzzle domains.
\end{abstract}

\section{Introduction}

Deep reinforcement learning (RL) has demonstrated remarkable success across various domains, including games \citep{Schrittwieser_2020, silver2017mastering, badia2020agent57}, robotics \citep{kalashnikov2018qtoptscalabledeepreinforcement, nagab2018learning}, and sequential decision-making tasks \citep{kim2021landmarkguided, czechowski2024subgoalsearchcomplexreasoning}. Despite these achievements, environments requiring complex, long-term decision-making remain a significant challenge. Hierarchical reinforcement learning (HRL) provides a promising avenue for addressing this issue by structuring decision-making across multiple levels of abstraction \citep{kim2021landmarkguided, SUTTON1999181, NIPS1992_d14220ee, dietterich1999hierarchical}. In HRL, a high-level policy, or "manager," decomposes a task into a series of subgoals, which are then executed by a lower-level "worker" policy. This hierarchical approach enables agents to leverage temporal abstraction, improving their ability to manage extended decision horizons and delayed rewards.

Several HRL methodologies have been proposed in recent years. One widely studied approach is option-based learning \citet{SUTTON1999181}, where a library of high-level skills (options) is developed, allowing an agent to operate at different temporal scales. The option-critic framework, for instance, learns both option policies and termination conditions in an end-to-end manner, eliminating the need for predefined subgoals \citep{bacon2016optioncriticarchitecture}. These options function as reusable skills, helping agents accelerate learning by avoiding repetitive low-level decision sequences. Another approach, feudal reinforcement learning, introduces hierarchical control, where a high-level policy sets abstract goals, while a lower-level policy refines them into actionable steps \citep{NIPS1992_d14220ee, vezhnevets2017feudalnetworkshierarchicalreinforcement}. Additionally, goal-conditioned HRL allows for dynamic goal selection, enabling an agent to reach various target states specified by the high-level controller. Techniques such as hindsight learning \citep{andrychowicz2018hindsightexperiencereplay}, which retrospectively labels past experiences as successful subgoal achievements, have further enhanced the effectiveness of goal-conditioned HRL in sparse-reward environments \citep{kim2021landmarkguided}. These advances—including options, feudal architectures, and goal-conditioned subpolicies—highlight the strengths of hierarchical decision-making in enabling agents to solve complex, long-horizon tasks.

The success of HRL largely depends on the identification of meaningful subgoals that guide decision-making. Recent research has focused on improving exploration by imposing constraints or guidance on high-level policies. For example, \citep{czechowski2024subgoalsearchcomplexreasoning} restrict the high-level policy to generate subgoals k steps ahead and demonstrates its effectiveness. Another effective strategy is to direct exploration toward promising landmark states—key states that play a crucial role in successful trajectories. \citep{kim2021landmarkguided} introduce HIGL (Hierarchical RL Guided by Landmarks), an approach that identifies and samples a diverse set of landmark states, steering the high-level policy toward them. By constructing a graph of visited landmarks and leveraging shortest-path planning, HIGL ensures more structured exploration compared to random wandering.

In this work, we propose HalfWeg, a hierarchical RL framework that follows a recursive top-down planning strategy. At each hierarchical level, subgoals are generated and passed down to lower levels for execution. These subgoals, or landmark states, are constrained to be reached within a fixed number of steps. A key feature of our framework is its ability to improve the hierarchy of policies in parallel by searching through sampled trajectories. This approach enables the model to learn the environment’s navigation dynamics without requiring handcrafted features or prior knowledge.

\begin{figure}[ht]
    \centering
    \includegraphics[width=0.32\textwidth]{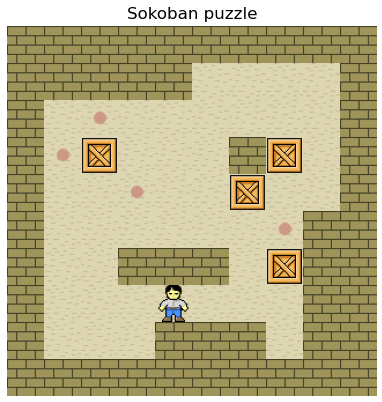}
    \caption{Sokoban example puzzle}
    \label{fig:fig4}
\end{figure}

To evaluate our approach, we experiment with the Sokoban \citep{sokoban_wiki} puzzle (a puzzle video game in which the player pushes boxes around in a warehouse, trying to get them to storage locations) and its variant, Boxoban \citep{boxobanlevels}, which constrains the number of boxes and level size. Sokoban presents a challenging planning problem, as irreversible mistakes can lead to unsolvable configurations. For training, we use the unfiltered level set, train our model on a training subset, and report all evaluations on previously unseen validation and test subsets.

Despite having significantly fewer parameters than contemporary models, our learned agents exhibit strong long-term planning capabilities in Sokoban, achieving performance comparable to state-of-the-art methods. Furthermore, we demonstrate that our trained policy effectively utilizes additional computation time, yielding substantially improved results when combined with search.

\section{Related work}

Model-free planning aims to implicitly learn planning by designing neural network architectures that approximate traditional planning algorithms \citep{guez2019investigationmodelfreeplanning}. Several works have explored this direction: \citet{farquhar2018treeqnatreecdifferentiabletreestructured} propose neural networks structured as search trees, \citet{guez2018learningsearchmctsnets} approximate Monte Carlo Tree Search (MCTS), \citet{tamar2017valueiterationnetworks} incorporate dynamic programming principles, and \citet{lehnert2024abetterplanningtransformers} develop architectures that approximate A* search. Following this approach, we train a hierarchy of policies, where each policy approximates search using trajectories sampled by the preceding lower-level policy. The first policy in this sequence acts as an approximation of exhaustive search, forming the foundation for subsequent policies.

Subgoal-based HRL methods have shown strong performance but remain highly sensitive to subgoal design \citep{dwiel2019hierarchicalpolicylearningsensitive}. The kSubS subgoal search algorithm \citep{czechowski2024subgoalsearchcomplexreasoning} generates subgoals k steps ahead and trains a low-level policy to navigate between them. Similarly, \citep{li2019hierarchicalreinforcementlearningadvantagebased} propose an HRL framework comprising high-level and low-level policies, where each high-level decision corresponds to k low-level execution steps. \citep{levy2019learningmultilevelhierarchieshindsight} introduce the Hierarchical Actor-Critic (HAC) framework, which learns a hierarchy of three policies by structuring the RL problem differently at each level. HAC assumes that lower-level policies are already optimal, allowing higher-level policies to be trained accordingly. While their hierarchical structure shares similarities with ours, we adopt a completely different policy training strategy.

\section{The HalfWeg framework for solving Sokoban}

The HalfWeg framework provides a structured approach for organizing and training a hierarchy of policies that recursively interact to solve navigation-based puzzles. In the case of Sokoban, solving the puzzle involves finding a sequence of actions that transforms an initial state into a goal state where all boxes are correctly positioned. Our method follows a top-down planning strategy, in which higher-level policies generate subgoals for lower levels, while the lowest level is responsible for executing primitive actions. To enable effective navigation between states, each policy must be capable of planning transitions from any given state to any target state. Therefore, rather than focusing solely on placing boxes in goal positions, we address a more general planning problem: understanding the dynamics of Sokoban puzzles and learning to compute optimal paths between any two puzzle states.

Our work builds directly on established principles of hierarchical reinforcement learning and subgoal-based planning, and demonstrates the following key attributes:

\begin{itemize}
    \item Recursive decomposition of goals into subgoals
    \item Automatic subgoal discovery without requiring additional domain knowledge
    \item A deep, 6-level hierarchical agent trained entirely from scratch to solve Sokoban
    \item Simplicity of function approximators — only two lightweight ResNet-based models are used across all six policies
    \item Joint training of all hierarchical levels
\end{itemize}

\subsection{Planning problem instance}

Instead of directly solving Sokoban, we redefine the problem as a more general planning task. Each instance of the planning problem is represented by the tuple $(u, v, b)$ where:

\begin{itemize}
    \item $u$ is the starting puzzle state, from which the agent begins its planning process.
    \item $v$ represents the target state, which serves as the objective of the planning task. The states $u$ and $v$ are not necessarily derived from the same initial puzzle configuration.
    \item $b$ is a directional control flag, where $b \in \{0,1\}$. If $b=0$, the objective is to create a plan that moves the agent as close as possible to the target state $v$. If $b=1$, the goal is reversed—the generated plan should move the agent as far as possible from $v$, effectively encouraging exploration.
\end{itemize}

In our framework, solving the original Sokoban puzzle is reformulated as solving planning instances of the form $(u, v, 0)$, where $u$ is the initial puzzle state and $v$ represents the desired goal state. To construct $v$, we ensure that all boxes are placed in their designated goal positions. However, the final position of the player in the solved state is unknown. To address this, we experiment with two approaches: (1) randomly assigning the player to an available empty cell or (2) placing the player in every possible empty cell and solving each resulting problem instance.

Sokoban presents a classic sparse reward problem, where the agent receives no intermediate feedback and only obtains a reward upon successfully solving the puzzle. In environments like Boxoban, this can require a large number of steps, making learning particularly challenging. Sparse rewards are widely recognized as one of the most difficult aspects of RL. To address this, we reformulate the reward function using a generalized planning approach, defining it as the distance between the target state $v$ and the actual state $v'$ reached after executing the proposed plan. This alternative reward structure provides denser and smoother feedback, making it more sensitive to incremental algorithmic improvements compared to the original sparse reward.

\subsection{Neural network models}

Our method utilizes two types of deep neural network models — both take a planning problem instance as input, but one outputs a sequence of actions while the other generates a landmark state:

\begin{itemize}
    \item Model Actions network $MA_\phi(u, v, b)$, parametrized by $\phi$, receives a planning problem instance and outputs a sequence of $d$ discrete actions.
    \item Model State network $MS_\theta(u, v, b, r)$, parameterized by $\theta$, takes the planning problem instance and recursion depth $r$ as input and predicts a landmark state $w$. This landmark state splits the original planning task — from $u$ to $v$ — into two smaller planning subproblems: navigating from $u$ to $w$ and then from $w$ to $v$. The recursion depth $r \in \{1 \dots R\}$ encodes the level of abstraction, where higher recursion levels $r=R$ focus on generating high-level subgoals, while lower levels $r=1$ are expected to solve the task within $2d$ actions.
\end{itemize}

\subsection{Recursive policies}

The purpose of our framework is to efficiently learn a hierarchy of policies $PL_0, \dots, PL_R$, where $R$ is a hyperparameter set by the user. Each policy $PL_i$ takes a planning problem instance $(u, v, b)$ as input and generates an action sequence of length $2^id$:

\begin{itemize}
    \item Level zero policy $PL_0$ directly calls the actions model: $PL_0(u, v, b) = MA(u, v, b)$; it returns a sequence with $d$ actions.
    \item Level $i \in \{1 \dots R\}$ policy $PL_i(u, v, b)$ follows a hierarchical planning approach:
        \begin{itemize}
            \item[$\diamond$]{Call the model $MS(u, v, b, i)$ to obtain sub-goal $w$.}
            \item[$\diamond$]{Call the underlying policy $PL_{i-1}(u, w, 0)$ to obtain a sequence of actions $a_1$ of length $2^{i-1}d$. The direction flag is set to 0, ensuring that the lower-level policy moves the agent as close as possible to $w$, irrespective of the original goal $v$ and direction flag $b$.}
            \item[$\diamond$]{Call the emulator starting from state $u$, execute actions $a_1$, and obtain the actual intermediate state $\hat{w}$.}
            \item[$\diamond$]{Call the underlying policy $PL_{i-1}(\hat{w}, v, b)$ to obtain a sequence of actions $a_2$ of length $2^{i-1}d$.}
            \item[$\diamond$]{Concatenate $a_1$ and $a_2$ to produce a final action plan of length $2^id$.}
        \end{itemize}
\end{itemize}

The top-level policy $PL_R$ is responsible for generating action sequences of length $2^Rd$. Increasing $R$ by 1 doubles both the computational effort required to compute the recursive call tree and the length of the generated plan, and allow planning over an exponentially larger space of possible plans: $A^{2^{R+1}d} = (A^{2^Rd})^2$. This exponential growth in the search space is key to improving long-term planning capabilities.

\section{Improving policies}

We view the collection of policies $PL_0, \dots, PL_R$ as an ensemble of dependent machine learning models. The goal of the HalfWeg framework is to improve policies so that they generate better plans. To do so, we make each policy an approximation of the search for a solution to a planning problem. The search process requires us to define a method for sampling a diverse set of action plans. We sample action plans using random goals, and in order to have a pool of random goals from which to sample, we include an exploration stage where agents (represented by policies) explore Sokoban puzzles. In summary, the HalfWeg training process consists of repeatedly executing the following steps (each of which will be described in detail later):


\begin{itemize}
    \item Generate gameplay trajectories by running agents in Sokoban levels sampled from the training set. Agents take random actions and pursue randomly sampled goal states to ensure diverse exploration.
    \item Sample planning problem instances $(u, v, b)$ from the experience replay buffer gathered during the previous step.
    \item Solve each planning instance using search guided by policies. Evaluate each plan based on the distance between its outcome and the target state $v$, and retain the best plan $a$ for training. The resulting training dataset contains tuples of the form $((u, v, b), a)$.
    \item Update all policies via gradient descent, using the collected training dataset.
\end{itemize}

\subsection{Self play and sampling of planning problems}

The training dataset consists of sampled planning problems and their corresponding solution plans $a$, proposed by the ensemble of policies. In the very first iteration, the policies do not produce meaningful plans, so we can only sample initial Sokoban puzzle states and states derived from them through random walks. As the policies improve, we must also sample states that lie farther from the initial random walk trajectories, since such states are qualitatively different. For example, until a puzzle is actually solved, we never observe a state where all goal cells are occupied by boxes.

To do so, each training iteration begins with a self-play exploration stage. We repeatedly sample random initial Sokoban states, and then either take a few random actions or call a randomly chosen policy $PL_i(u, v, b)$, where $u$ is the current Sokoban state, $v$ is sampled from previously seen states stored in the experience replay buffer, and $b \in \{0,1\}$ is randomly chosen. In other words, the agent either performs random moves or follows a plan that moves it toward or away from a random goal. When $b=0$, the policy is expected to generate a plan that leads toward the "center" of the well-explored state space, as target states $v$ are more likely to be sampled from frequently visited regions. In contrast, when $b=1$, the policy is encouraged to move away from well-explored areas, promoting broader exploration of the environment.

\subsection{Solving planning problems via search}

In this step, we take a sampled planning tuple $(u, v, b)$ and call each policy to propose its solution using search. Search allows us to find better solutions compared to a single forward pass of a policy, so we invest additional computational effort at this stage to identify higher-quality plans.

The base policy $PL_0$ relies on exhaustive search to identify the optimal plan of length $d$. With an action space of size $A$, it systematically evaluates all $A^d$ possible action sequences and returns the one that yields the best result according to the distance metric.

Each policy $PL_i, i > 0$ is designed to output plans of length $2^id$. It implements search using the following procedure:

\begin{itemize}
    \item Sample $N_{DSS}$ (set to 100 in our experiments) random intermediate subgoal states $w_j$ from experience replay, along with random direction flags $b_j \in \{0,1\}$.
    \item Call the underlying policy $PL_{i-1}(u, w_j, b_j)$ to obtain a sequence of actions $a_{1,j}$ of length $2^{i-1}d$.
    \item Call the emulator starting from state $u$ using actions $a_{1,j}$ to obtain a list of $N_{DSS}$ states $\hat{u}_j$.
    \item Call the underlying policy $PL_{i-1}(\hat{u}_j, v, b)$ to obtain a sequence of actions $a_{2,j}$ of length $2^{i-1}d$.
    \item Call the emulator starting from states $\hat{u}_j$ using actions $a_{2,j}$ to obtain a list of $N_{DSS}$ states $\hat{v}_j$.
    \item Find an index $j^{*}$ that optimizes the distance metric between $\hat{v}_j$ and $v$ based on the original direction flag $b$.
    \item Return the concatenation of $a_{1,j^{*}}$ and $a_{2,j^{*}}$ as the final plan of length $2^id$ for a policy $PL_i$.
\end{itemize}

Each planning problem is evaluated by all policies. Each policy $PL_i$ proposes a plan of length $2^id$, computed using two-leg search with its underlying policy (or exhaustive search in the case of $PL_0$). If we were to use each policy’s plan to train it directly, then each $PL_i$ would become an approximation of the search of the policy one level below it. However, in our framework, we include in the training dataset only the best plan $a_{\hat{i}}$ among those proposed by all policies. As a result, each policy $PL_i$ becomes an approximation not just of its own underlying policy's search, but the best result from a collective search over the entire hierarchy of policies (including itself, but excluding the final policy $PL_R$).

\subsection{Train planning models}

The result of the search step described above is a training dataset consisting of planning problems $(u, v, b)$ and their corresponding plans $a$, found by the search process. The plans $a$ may vary in length, as they are not necessarily generated by the same policy. Each row in the dataset is then used to perform stochastic gradient descent updates on the two neural network models, $MA$ and $MS$, according to the following logic:

\begin{itemize}
    \item Call the emulator starting from state $u$ with actions $a$ to obtain a sequence of states $u_i$ along the trajectory. Each state $u_i$ corresponds to executing the prefix $a_{1 \dots i}$ starting from $u$.
    \item Model $MA_\phi(u, v, b)$ outputs a sequence of actions $a'$ of length d. To train this model we perform training iteration using $(u, v, b)$ as input and the prefix $a_{1 \dots d}$ as the target. We optimize the cross-entropy loss between predicted $a'$ and actual prefix $a_{1 \dots d}$
    \item We also include a refinement example $(u, u_d, b=0)$ with target $a_{1 \dots d}$ in the training set for the $MA$ model. This refinement row helps the model learn to navigate when the goal state lies within $d$ steps. Our ablation study (see below) confirms that using such examples improves training.
    \item Model $MS_\phi(u, v, b, r)$ outputs an intermediate state $w$. For each $r \in (1 \dots R)$, we use $(u, v, b, r)$ as input and the state $u_{2^{r-1}d}$ as the target. If the action sequence $a$ is shorter than $2^{r-1}d$ then we use the last final state $u_{|a|}$ as the target. The training loss is the mean squared error between the predicted state $w$ and the target state.
    \item Additionally, we add a refinement row for the $MS$ model with input $(u, u_{|a|}, b=0)$ and target $u_{2^{r-1}d}$.
\end{itemize}

\section{Experiments}

To demonstrate our method's performance, we present experimental results across three sets of experiments: (1) the Boxoban environment; (2) smaller Sokoban procedurally generated levels, used to illustrate training dynamics and perform an ablation study; and (3) a generalization experiment designed to evaluate the agent’s performance on levels generated out of the training distribution.

\subsection{Boxoban performance}

Boxoban is a standardized variant of Sokoban featuring a fixed 10x10 grid and four boxes. The environment is publicly available \citep{boxobanlevels} and includes a consistent train/validation/test split, enabling fair comparisons across different methods.

We trained the HalfWeg agent with the $MA$ model containing 383,000 free parameters and the $MS$ model containing 306,000 free parameters. The detailed architecture and training procedure are presented in the supplementary materials. Figure \ref{fig:fig1} shows the training progress of the Boxoban policies. Training was conducted on the Boxoban training set.

\begin{figure}[ht]
    \centering
    \includegraphics[width=0.77\textwidth]{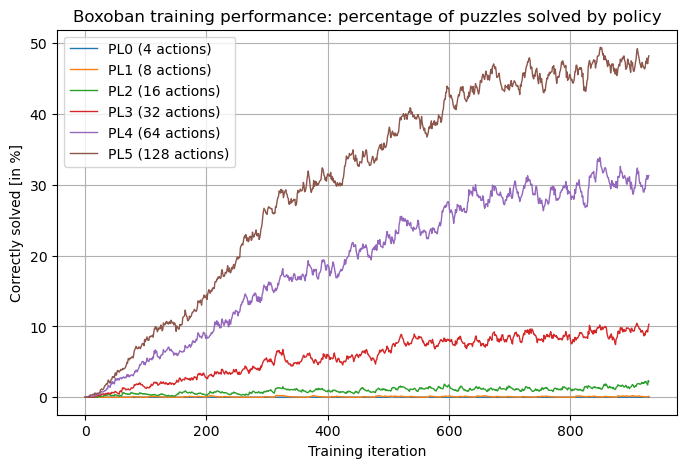}
    \caption{Training the HalfWeg algorithm on Boxoban using models with a total of 689,496 free parameters and a hierarchy of 6 policies. The Y-axis shows the percentage of solved Boxoban validation test instances (unseen during training) with 36 targets}
    \label{fig:fig1}
\end{figure}

Table \ref{tab:experiment_logs} shows the effectiveness of the trained policies and how the number of correctly solved puzzles improves with the use of search. The columns in this table are:

\begin{itemize}
    \item Solved – Percentage of solved Boxoban test instances.
    \item Targets – A value of 1 means only one target state is created by placing the player at a random empty position. Higher values indicate that the player is placed in all possible empty positions, and the agent is asked to solve at least one of them.
    \item Searches – A value of '–' indicates no search is used, the hierarchy of models is called exactly once for every target. Any other value indicates the use of two-leg search, as described above, with the corresponding number of random goals. The total number of policy calls is equal to the number of targets multiplied by the number of searches.
    \item Solution length – Average length of the plans across successfully solved puzzles.
\end{itemize}

\begin{table}[ht]
\centering
\begin{tabular}{|c|c|c|c|c|}
\hline
\textbf{Policy} & \textbf{Solved \%} & \textbf{Targets} & \textbf{Searches} & \textbf{HalfWeg solution length} \\
\hline
$PL_2$ & 0.4 & 1 & - & 14.5 \\
$PL_3$ & 3.7 & 1 & - & 21.4 \\
$PL_4$ & 8.3 & 1 & - & 37.0 \\
$PL_5$ & 12.9 & 1 & - & 65.6 \\
$PL_2$ & 0.6 & 36 & - & 14.2 \\
$PL_3$ & 10.6 & 36 & - & 25.1 \\
$PL_4$ & 31.3 & 36 & - & 50.1 \\
$PL_5$ & 46 & 36 & - & 102.8 \\
$PL_5$ & 79.4 & 36 & 100 & 189.4 \\
$PL_5$ & 90.2 & 36 & 1000 & 214.7 \\
\hline
\end{tabular}
\caption{Summary of trained agent performance on the Boxoban puzzle}
\label{tab:experiment_logs}
\end{table}

Notably, we observe that the final policy $PL_5$ is able to solve 13\% of unseen Boxoban puzzles with a single forward pass (i.e., one tree call), with an average solution length of 65 steps—demonstrating the agent’s ability for long-term planning. When increasing the number of targets, we observe significant improvements. By evaluating all possible player positions in the target state and using only one call to policy $PL_5$ per target, the agent solves 46\% of the puzzles, with an average plan length of 102.8 steps. This indicates that the trained policy is capable of identifying meaningful landmark states and generating complete plans in a top-down, recursive fashion. Further gains are observed when augmenting $PL_5$ with search: with 36,000 total policy calls (as shown in the last row of the table), the agent solves over 90\% of the test puzzles, producing solutions with an average length of 214.7 steps. Due to computational constraints, we did not experiment with larger model sizes for Boxoban. However, our ablation study on Sokoban demonstrates that increasing model size significantly improves planning performance.

\subsection{Training and ablation study on generated smaller Sokoban set}

In this set of experiments, we use Sokoban levels of size 6x6 with 3 boxes. The levels are generated on the fly using a custom level generator.

\begin{figure}[ht]
    \centering
    \includegraphics[width=0.49\textwidth]{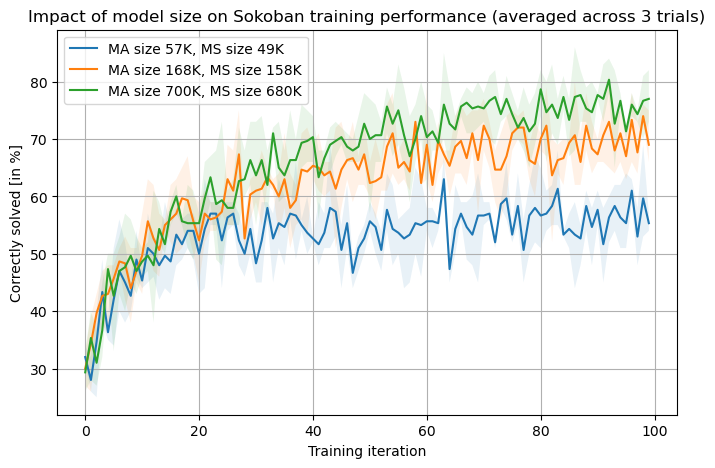}
    \hfill
    \includegraphics[width=0.49\textwidth]{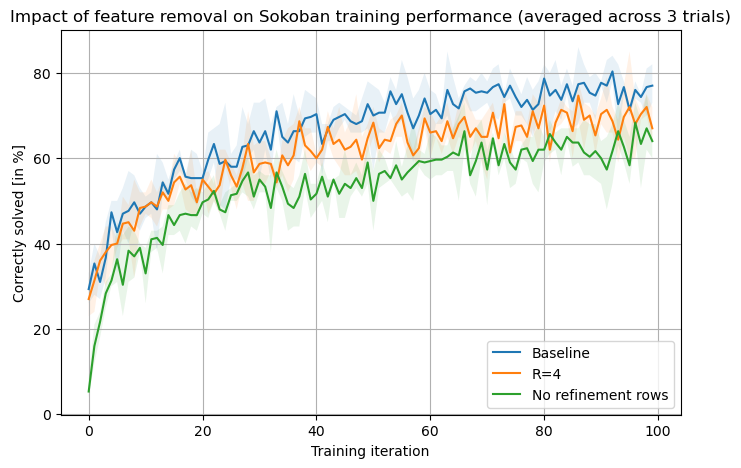}
    \caption{Impact of model sizes ($MA$ and $MS$) and the effect of feature removal. Both plots show training curves for Sokoban 6x6 with 3 boxes.}
    \label{fig:fig2}
\end{figure}

Figure \ref{fig:fig2} (left) shows the impact of model sizes on the training process. We observe that increasing model size improves training performance, while the smallest model exhibits an early plateau. On the right, we show the effect of various feature removals:

\begin{itemize}
    \item Baseline corresponds to the run with an $MA$ model containing 700,000 free parameters and an $MS$ model with 680,000 parameters. This setup uses 6 policies, with the top-level policy being $PL_5$ (i.e., $R=5$).
    \item $R=4$ represents a run using only 5 policies, with $PL_4$ as the highest level. We observe that the $R=5$ setting performs better than $R=4$, but not drastically so. For this reason, we did not experiment with $R=6$.
    \item "No refinement rows" denotes a run where all refinement rows were removed from the training dataset, as described above.
\end{itemize}

\subsection{Generalization experiment}

\begin{figure}[ht]
    \centering
    \includegraphics[width=0.85\textwidth]{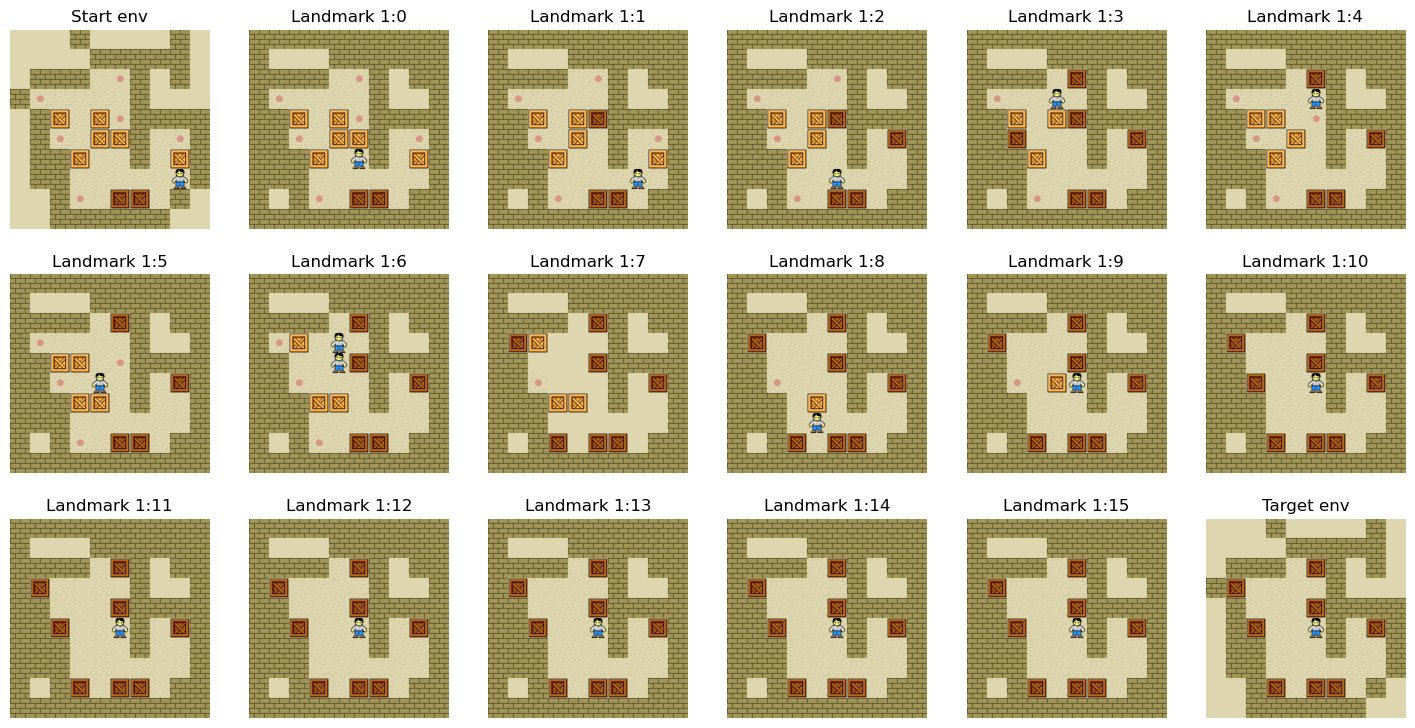}
    \caption{Example of envisioned landmark sequences for a generated Sokoban puzzle with 8 boxes}
    \label{fig:fig_sokoban_8_boxes}
\end{figure}

We evaluated the trained Boxoban (10x10 level with 4 boxes) agent on levels generated by our custom Sokoban level generator. To assess generalization, we evaluated the trained policy $PL_5$ on levels generated by our tool, which produces puzzles of the same size (10x10) but with varying numbers of boxes. In this experiment, we generated target states by placing all boxes in their goal positions and trying all possible player positions. The policy was invoked once per target state, and no search was used — i.e., the policy output was used directly. Table \ref{tab:generalization_table} presents the performance of policy $PL_5$ on these levels. Despite being trained exclusively on Boxoban levels with exactly 4 boxes, the agent demonstrates the ability to generalize and solve levels with different numbers of boxes. Figure \ref{fig:fig_sokoban_8_boxes} shows an example of a generated landmarks for Sokoban level with 8 boxes.

\begin{table}[ht]
\centering
\begin{tabular}{|c|c|c|c|}
\hline
\textbf{Boxes \%} & \textbf{Solved \%} & \textbf{Generator solution length} & \textbf{HalfWeg solution length} \\
\hline
1 & 93.8 & 12.3 & 115.0 \\
2 & 76.8 & 22.0 & 105.4 \\
3 & 73.3 & 22.7 & 103.3 \\
4 & 76.7 & 21.9 & 99.98 \\
5 & 79.5 & 20.4 & 100.95 \\
6 & 79.1 & 19.1 & 97.15 \\
7 & 82.1 & 18.0 & 98.2 \\
8 & 79.8 & 17.7 & 96.4 \\
9 & 79.4 & 17.2 & 89.4 \\
\hline
\end{tabular}
\caption{Summary of trained agent performance on generated puzzles with various numbers of boxes.}
\label{tab:generalization_table}
\end{table}

\section{Supplementary}

\subsection{Models details}

Both models $MA$ and $MS$ take as input a generalized planning problem tuple $(u, v, b)$. The states $u$ and $v$ are encoded as tensors of shape $(B, C, H, W)$, where $B$ is the batch size, $C$ is the number of channels used to represent each cell in the puzzle (for Sokoban, $C=4$, corresponding to the player, wall, box, and goal), and $H, W$ denote the height and width of the board, respectively.

The length of the action list for $MA$ is $d$ (We use $d=4$ for every model in this work.), and to allow the agent to produce plans whose effective length is not divisible by $d$, we introduce an additional action value that signals the end of the plan. Subsequent actions are ignored after this stop signal. In the Sokoban environment, the action space consists of 5 values: 0 (up), 1 (right), 2 (left), 3 (down), and 4 (stop). As a result, the output of the $MA$ model is a tensor of shape 
$(B, d, 5)$, where $B$ is the batch size, $d$ is the maximum number of actions in the plan, and 5 corresponds to the probability distribution over possible moves at each step.

The $MA$ and $MS$ models share a similar architecture: both begin with a 3x3 convolutional layer, followed by a series of ResNet blocks. The $MA$ model ends with a linear layer that transforms the convolutional output into a tensor of shape $(B, d, 5)$. In contrast, the $MS$ model ends with a convolutional layer that produces a tensor of shape $(B,C,H,W)$, matching the shape of the puzzle state representation.

In this work, we use the following neural network models:

\begin{itemize}
    \item Sokoban 6x6 smallest model with 48 filters and 1 ResNet block: $MA$ has 57263 trainable free parameters; $MS$ has 49060 parameters.
    \item Sokoban 6x6 medium model with 64 filters and 2 ResNet blocks: $MA$ has 168719 trainable free parameters; $MS$ has 157828 parameters.
    \item Sokoban 6x6 large model with 96 filters and 4 ResNet blocks: $MA$ has 696143 trainable free parameters; $MS$ has 679876 parameters.
    \item Boxoban 10x10 main model with 64 filters and 4 ResNet blocks: $MA$ has 383124 trainable free parameters; $MS$ has 306372 parameters.
\end{itemize}

\subsection{Model planning showcase}

\begin{figure}[ht]
    \centering
    \includegraphics[width=0.85\textwidth]{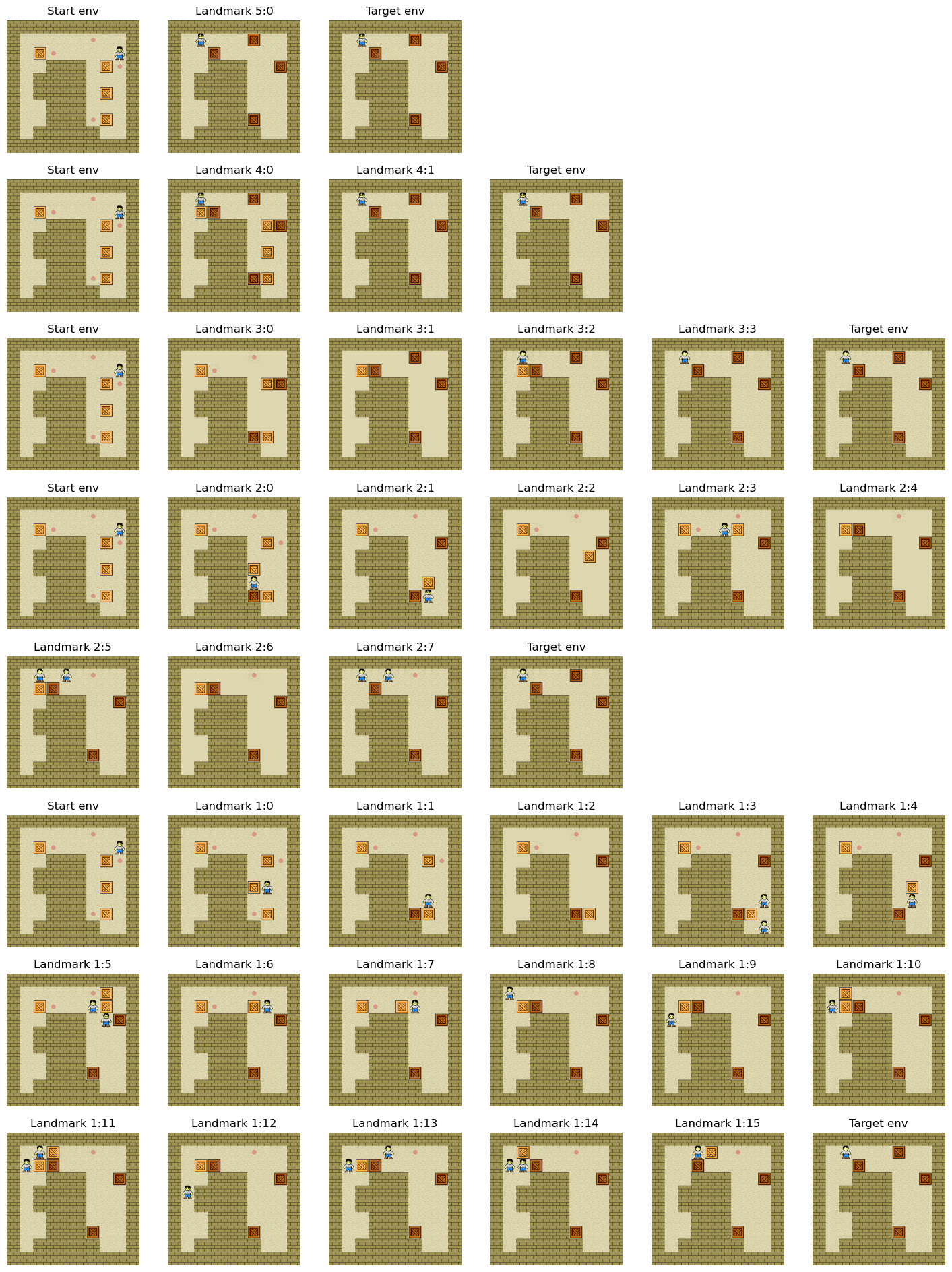}
    \caption{Tree of landmarks envisioned by the Boxoban model (policy $PL_5$) to solve a Boxoban level.}
    \label{fig:fig3}
\end{figure}

Figure \ref{fig:fig3} shows an example of landmarks generated by the $PL_5$ Boxoban model for each call to the $MS$ model within the recursive call tree. The first image at each recursion level is the starting puzzle state, and the last image is the target state used for planning. The intermediate images represent the envisioned landmark states. Notably, the landmark sequences are easy to interpret at the two lowest levels (sequences Landmark 2:* and Landmark 1:*). In these layers, we can clearly observe how the model plans to move each box, one by one.

\newpage

\bibliography{main}

\begin{thebibliography}{23}
\providecommand{\natexlab}[1]{#1}
\providecommand{\url}[1]{\texttt{#1}}
\expandafter\ifx\csname urlstyle\endcsname\relax
  \providecommand{\doi}[1]{doi: #1}\else
  \providecommand{\doi}{doi: \begingroup \urlstyle{rm}\Url}\fi

\bibitem[Andrychowicz et~al.(2018)Andrychowicz, Wolski, Ray, Schneider, Fong, Welinder, McGrew, Tobin, Abbeel, and Zaremba]{andrychowicz2018hindsightexperiencereplay}
Marcin Andrychowicz, Filip Wolski, Alex Ray, Jonas Schneider, Rachel Fong, Peter Welinder, Bob McGrew, Josh Tobin, Pieter Abbeel, and Wojciech Zaremba.
\newblock Hindsight experience replay, 2018.
\newblock URL \url{https://arxiv.org/abs/1707.01495}.

\bibitem[Bacon et~al.(2016)Bacon, Harb, and Precup]{bacon2016optioncriticarchitecture}
Pierre-Luc Bacon, Jean Harb, and Doina Precup.
\newblock The option-critic architecture, 2016.
\newblock URL \url{https://arxiv.org/abs/1609.05140}.

\bibitem[Badia et~al.(2020)Badia, Piot, Kapturowski, Sprechmann, Vitvitskyi, Guo, and Blundell]{badia2020agent57}
Adrià~Puigdomènech Badia, Bilal Piot, Steven Kapturowski, Pablo Sprechmann, Alex Vitvitskyi, Daniel Guo, and Charles Blundell.
\newblock Agent57: Outperforming the atari human benchmark, 2020.

\bibitem[Czechowski et~al.(2024)Czechowski, Odrzygóźdź, Zbysiński, Zawalski, Olejnik, Wu, Łukasz Kuciński, and Miłoś]{czechowski2024subgoalsearchcomplexreasoning}
Konrad Czechowski, Tomasz Odrzygóźdź, Marek Zbysiński, Michał Zawalski, Krzysztof Olejnik, Yuhuai Wu, Łukasz Kuciński, and Piotr Miłoś.
\newblock Subgoal search for complex reasoning tasks, 2024.
\newblock URL \url{https://arxiv.org/abs/2108.11204}.

\bibitem[Dayan \& Hinton(1992)Dayan and Hinton]{NIPS1992_d14220ee}
Peter Dayan and Geoffrey~E Hinton.
\newblock Feudal reinforcement learning.
\newblock In S.~Hanson, J.~Cowan, and C.~Giles (eds.), \emph{Advances in Neural Information Processing Systems}, volume~5. Morgan-Kaufmann, 1992.
\newblock URL \url{https://proceedings.neurips.cc/paper_files/paper/1992/file/d14220ee66aeec73c49038385428ec4c-Paper.pdf}.

\bibitem[Dietterich(1999)]{dietterich1999hierarchical}
Thomas~G. Dietterich.
\newblock Hierarchical reinforcement learning with the maxq value function decomposition, 1999.

\bibitem[Dwiel et~al.(2019)Dwiel, Candadai, Phielipp, and Bansal]{dwiel2019hierarchicalpolicylearningsensitive}
Zach Dwiel, Madhavun Candadai, Mariano Phielipp, and Arjun~K. Bansal.
\newblock Hierarchical policy learning is sensitive to goal space design, 2019.
\newblock URL \url{https://arxiv.org/abs/1905.01537}.

\bibitem[Farquhar et~al.(2018)Farquhar, Rocktäschel, Igl, and Whiteson]{farquhar2018treeqnatreecdifferentiabletreestructured}
Gregory Farquhar, Tim Rocktäschel, Maximilian Igl, and Shimon Whiteson.
\newblock Treeqn and atreec: Differentiable tree-structured models for deep reinforcement learning, 2018.
\newblock URL \url{https://arxiv.org/abs/1710.11417}.

\bibitem[Guez et~al.(2018{\natexlab{a}})Guez, Mirza, Gregor, Kabra, Racaniere, Weber, Raposo, Santoro, Orseau, Eccles, Wayne, Silver, Lillicrap, and Valdes]{boxobanlevels}
Arthur Guez, Mehdi Mirza, Karol Gregor, Rishabh Kabra, Sebastien Racaniere, Theophane Weber, David Raposo, Adam Santoro, Laurent Orseau, Tom Eccles, Greg Wayne, David Silver, Timothy Lillicrap, and Victor Valdes.
\newblock An investigation of model-free planning: boxoban levels.
\newblock https://github.com/deepmind/boxoban-levels/, 2018{\natexlab{a}}.

\bibitem[Guez et~al.(2018{\natexlab{b}})Guez, Weber, Antonoglou, Simonyan, Vinyals, Wierstra, Munos, and Silver]{guez2018learningsearchmctsnets}
Arthur Guez, Théophane Weber, Ioannis Antonoglou, Karen Simonyan, Oriol Vinyals, Daan Wierstra, Rémi Munos, and David Silver.
\newblock Learning to search with mctsnets, 2018{\natexlab{b}}.
\newblock URL \url{https://arxiv.org/abs/1802.04697}.

\bibitem[Guez et~al.(2019)Guez, Mirza, Gregor, Kabra, Racanière, Weber, Raposo, Santoro, Orseau, Eccles, Wayne, Silver, and Lillicrap]{guez2019investigationmodelfreeplanning}
Arthur Guez, Mehdi Mirza, Karol Gregor, Rishabh Kabra, Sébastien Racanière, Théophane Weber, David Raposo, Adam Santoro, Laurent Orseau, Tom Eccles, Greg Wayne, David Silver, and Timothy Lillicrap.
\newblock An investigation of model-free planning, 2019.
\newblock URL \url{https://arxiv.org/abs/1901.03559}.

\bibitem[Kalashnikov et~al.(2018)Kalashnikov, Irpan, Pastor, Ibarz, Herzog, Jang, Quillen, Holly, Kalakrishnan, Vanhoucke, and Levine]{kalashnikov2018qtoptscalabledeepreinforcement}
Dmitry Kalashnikov, Alex Irpan, Peter Pastor, Julian Ibarz, Alexander Herzog, Eric Jang, Deirdre Quillen, Ethan Holly, Mrinal Kalakrishnan, Vincent Vanhoucke, and Sergey Levine.
\newblock Qt-opt: Scalable deep reinforcement learning for vision-based robotic manipulation, 2018.
\newblock URL \url{https://arxiv.org/abs/1806.10293}.

\bibitem[Kim et~al.(2021)Kim, Seo, and Shin]{kim2021landmarkguided}
Junsu Kim, Younggyo Seo, and Jinwoo Shin.
\newblock Landmark-guided subgoal generation in hierarchical reinforcement learning, 2021.

\bibitem[Lehnert et~al.(2024)Lehnert, Sukhbaatar, Su, Zheng, Mcvay, Rabbat, and Tian]{lehnert2024abetterplanningtransformers}
Lucas Lehnert, Sainbayar Sukhbaatar, DiJia Su, Qinqing Zheng, Paul Mcvay, Michael Rabbat, and Yuandong Tian.
\newblock Beyond a*: Better planning with transformers via search dynamics bootstrapping, 2024.
\newblock URL \url{https://arxiv.org/abs/2402.14083}.

\bibitem[Levy et~al.(2019)Levy, Konidaris, Platt, and Saenko]{levy2019learningmultilevelhierarchieshindsight}
Andrew Levy, George Konidaris, Robert Platt, and Kate Saenko.
\newblock Learning multi-level hierarchies with hindsight, 2019.
\newblock URL \url{https://arxiv.org/abs/1712.00948}.

\bibitem[Li et~al.(2019)Li, Wang, Tang, and Zhang]{li2019hierarchicalreinforcementlearningadvantagebased}
Siyuan Li, Rui Wang, Minxue Tang, and Chongjie Zhang.
\newblock Hierarchical reinforcement learning with advantage-based auxiliary rewards, 2019.
\newblock URL \url{https://arxiv.org/abs/1910.04450}.

\bibitem[Nagabandi et~al.(2018)Nagabandi, Clavera, Liu, Fearing, Abbeel, Levine, and Finn]{nagab2018learning}
Anusha Nagabandi, Ignasi Clavera, Simin Liu, Ronald~S. Fearing, Pieter Abbeel, Sergey Levine, and Chelsea Finn.
\newblock Learning to adapt in dynamic, real-world environments through meta-reinforcement learning, 2018.

\bibitem[Schrittwieser et~al.(2020)Schrittwieser, Antonoglou, Hubert, Simonyan, Sifre, Schmitt, Guez, Lockhart, Hassabis, Graepel, Lillicrap, and Silver]{Schrittwieser_2020}
Julian Schrittwieser, Ioannis Antonoglou, Thomas Hubert, Karen Simonyan, Laurent Sifre, Simon Schmitt, Arthur Guez, Edward Lockhart, Demis Hassabis, Thore Graepel, Timothy Lillicrap, and David Silver.
\newblock Mastering atari, go, chess and shogi by planning with a learned model.
\newblock \emph{Nature}, 588\penalty0 (7839):\penalty0 604–609, December 2020.
\newblock ISSN 1476-4687.
\newblock \doi{10.1038/s41586-020-03051-4}.
\newblock URL \url{http://dx.doi.org/10.1038/s41586-020-03051-4}.

\bibitem[Silver et~al.(2017)Silver, Hubert, Schrittwieser, Antonoglou, Lai, Guez, Lanctot, Sifre, Kumaran, Graepel, Lillicrap, Simonyan, and Hassabis]{silver2017mastering}
David Silver, Thomas Hubert, Julian Schrittwieser, Ioannis Antonoglou, Matthew Lai, Arthur Guez, Marc Lanctot, Laurent Sifre, Dharshan Kumaran, Thore Graepel, Timothy Lillicrap, Karen Simonyan, and Demis Hassabis.
\newblock Mastering chess and shogi by self-play with a general reinforcement learning algorithm, 2017.

\bibitem[Sutton et~al.(1999)Sutton, Precup, and Singh]{SUTTON1999181}
Richard~S. Sutton, Doina Precup, and Satinder Singh.
\newblock Between mdps and semi-mdps: A framework for temporal abstraction in reinforcement learning.
\newblock \emph{Artificial Intelligence}, 112\penalty0 (1):\penalty0 181--211, 1999.
\newblock ISSN 0004-3702.
\newblock \doi{https://doi.org/10.1016/S0004-3702(99)00052-1}.
\newblock URL \url{https://www.sciencedirect.com/science/article/pii/S0004370299000521}.

\bibitem[Tamar et~al.(2017)Tamar, Wu, Thomas, Levine, and Abbeel]{tamar2017valueiterationnetworks}
Aviv Tamar, Yi~Wu, Garrett Thomas, Sergey Levine, and Pieter Abbeel.
\newblock Value iteration networks, 2017.
\newblock URL \url{https://arxiv.org/abs/1602.02867}.

\bibitem[Vezhnevets et~al.(2017)Vezhnevets, Osindero, Schaul, Heess, Jaderberg, Silver, and Kavukcuoglu]{vezhnevets2017feudalnetworkshierarchicalreinforcement}
Alexander~Sasha Vezhnevets, Simon Osindero, Tom Schaul, Nicolas Heess, Max Jaderberg, David Silver, and Koray Kavukcuoglu.
\newblock Feudal networks for hierarchical reinforcement learning, 2017.
\newblock URL \url{https://arxiv.org/abs/1703.01161}.

\bibitem[Wikipedia(2024)]{sokoban_wiki}
Wikipedia.
\newblock {Sokoban} --- {W}ikipedia{,} the free encyclopedia.
\newblock \url{https://en.wikipedia.org/wiki/Sokoban}, 2024.

\end{thebibliography}
\bibliographystyle{tmlr}

\end{document}